\documentclass[letterpaper, 10 pt, conference]{ieeeconf}
\usepackage[pdftex]{graphicx}
\usepackage{tabularx}
\usepackage{colortbl}
\usepackage{hhline}

\usepackage{subcaption}
\usepackage{amsmath}
\usepackage{mathtools}

\usepackage{graphicx}
\usepackage[colorinlistoftodos]{todonotes}

\usepackage{blindtext}

\title{\LARGE \bf
 Adapting Everyday Manipulation Skills to Varied Scenarios
}

\IEEEoverridecommandlockouts

\author{Pawe\l{}  Gajewski$^{1}$, Paulo Ferreira$^{3}$, Georg Bartels$^{4}$, Chaozheng Wang$^{2}$, Frank Guerin$^{2}$, Bipin Indurkhya$^{1}$, \\ Michael Beetz$^{4}$ and Bart\l{}omiej \'Snie\.zy\'nski$^{1}$    
\thanks{$^{1}$AGH University of Science and Technology,
al. Mickiewicza 30, 30-059 Krakow, Poland
        {\tt\small pawel.gajewski@agh.edu.pl}}%
\thanks{$^{2}$Department of Computing Science, University of Aberdeen,
King's College, AB24 3UE, Aberdeen, Scotland}%
\thanks{$^{3}$School of Computer Science, University of Birmingham,
B15 2TT, Birmingham, England}%
\thanks{$^{4}$Universit\"at Bremen,
Am Fallturm 1, 28359 Bremen, Germany}%
        }

\begin{document}

\maketitle
\thispagestyle{empty}
\pagestyle{empty}

\noindent \begin{abstract}
\noindent We address the problem of executing tool-using  manipulation skills in  scenarios where the objects to be used may vary. We assume that point clouds of the tool and target object can be obtained, but no interpretation or further knowledge about these objects is provided. The system must interpret the point clouds and decide how to use the tool to complete a manipulation task with a target object; this means it must adjust motion trajectories appropriately to complete the task.
We tackle three everyday manipulations: scraping material from a tool into a container, cutting, and scooping from a container.
Our solution encodes these manipulation skills in a generic way, with parameters that can be filled in at run-time via queries to  a robot perception module; the perception module abstracts the functional parts of the tool and  extracts key parameters that are needed for the task.
The approach is evaluated in simulation and with selected examples on  a PR2 robot.

\end{abstract}

\section{INTRODUCTION}


\noindent Service robots are expected to be able to perform  everyday manipulation tasks in human environments in the not too distant future \cite{Ersen2017}.
We focus on common everyday manipulation skills with tools, such as cutting, scooping and scraping. Such tasks need to be executed across varied scenarios in order to complete common everyday domestic tasks. 
The challenge is to make robot manipulation skills which can adapt to the variations in open environments.
For example, there may be variations in the tools available, in the substances to be manipulated, in the target objects, and in their layout.
 We want to encode a manipulation skill in a generic way for our robot, so that if it encounters a new tool and target object, it can appropriately adapt its motion and complete the task.

For example consider the manipulation of scraping a sticky substance from a hand-held domestic tool into some container. We would like to endow the robot with this skill, coded in a sufficiently generic way that it could automatically adapt to scenarios such as depicted in Fig. 1: 1)
 peanut butter on a knife can be put back in a jar of peanut butter by scraping the knife on the inner edge of the jar opening; 2)
 butter  stuck to a spatula can be put in a frying pan by scraping on the inner edge of the pan.
 
A key feature of our problem is that a domestic robot should be able to deal with new objects autonomously.
Ideally we do not want to rely on a human designer to label parts of each object or indicate where to grasp and how to orient the object. We only assume that the robot can obtain point clouds of the objects. Also the robot has not been trained to manipulate these particular objects in advance. The robot must generate appropriate trajectories based on the information it can extract from perception.

\begin{figure}[t!]
	\centering
	\includegraphics[width=8cm]{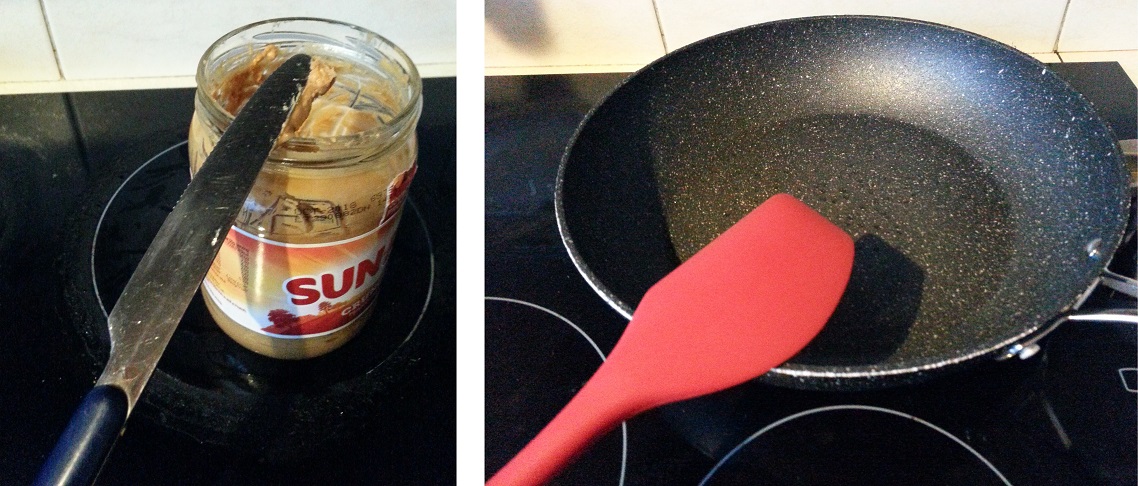}
	\caption{
	We want to achieve robust manipulation skills that can adapt to different tools and target objects, e.g. a generic `scrape' skill can: 1) scrape excess peanut butter from a knife into a jar (left) or 2) scrape butter into a frying pan (right).
}\label{jarpan}
\vspace{-0.5cm}
\end{figure}

This ability to adapt to varied scenarios (varied tools and target objects in our case) is one of the key abilities that is required to allow robots to function in more everyday human environments, and currently is limiting their deployment in domestic settings.
It is an interesting challenge because it seems to be so easy for humans (even very young humans) to outperform existing robot systems on many basic household manipulation tasks.
 Tackling this problem in a task-general way may shed some light on the reasons for the gap between humans and robots in everyday tasks. For example it is suggested that it may require an element of creativity \cite{fitz2017}.

The problem of adapting to varied scenarios is challenging because we need to connect information at high and low levels of abstraction, and also between  perceptual and motion control abilities. This can be illustrated with the example of adapting a scooping skill to work with a new tool: We have prior high-level knowledge that the suitable tool will have a concave bowl-like part, and some handle. This needs to be connected to the lower level  3D vision to find suitable parts in a suitable relationship which could constitute these components. 
Next we need to extract key parameters from the 3D vision that can be used to adjust the motion trajectories appropriately. For example to raise the tool higher above the container if this tool has a particularly long handle, and to maintain an appropriate orientation of the bowl-shaped scooping part while a substance is transported in it.
Not many works attempt to integrate in one system the required perception and motion components, and levels of abstraction.

To solve this problem it is critical to select a suitable representation for coding the manipulation skill. Our representation describes motion phases in terms of constraints among  key features of the tool and target objects, where these features can be grounded in perception by a robot vision system. Constraints are distances between key points on the objects, or angles between objects. 
Our system integrates two existing recent research works, one in robot vision \cite{AbelhaArXiv2017} and one in robot motion \cite{Tenorth2014,DBLP:conf/iros/FangBB16}. 
We rely on a hand coded generic task description for the manipulation skill, which represents the main steps involved in the motion, in terms of constraints that the motion solver will attempt to satisfy.
In this way, the actual motion trajectory for a particular tool and target object  is made on the spot.

In terms of the general approach of linking key features of objects to a constraint-based specification for motor control, the most closely related work to ours is Tenorth et al. \cite{Tenorth2014} but we make the following extra contributions:
\begin{itemize}
\item Their work requires CAD models, while we fit superquadrics to point clouds of unknown objects; this allows a robot to be more robust in an open environment, facing unexpected objects.
\item Their work had a hard-coded interpretation of objects for pouring; we learn affordances through simulated trials of tools \cite{AbelhaArXiv2017}, thereby acquiring a deeper semantic grounding for our task skills. This is a significant step towards more cognitive robots that form their own interpretation of the world rather than relying on the pre-formed interpretation of a human designer.
\item Their work only considers pouring and we go beyond, having three tasks: scraping, cutting and scooping. This is important to show the general nature of our system: the way in which we connect visual interpretation of objects to adaptation of motions  can be applied to many tasks. This contrasts with many leading robotics works which focus on a single task \cite{leidner2017phd,Dogar-2013-7755}.
\end{itemize}

In summary our contribution is an integrated system, 
with ROS implementation\footnote{https://github.com/lubiluk/skill\_transfer},
that recognises how to use a tool from its point cloud,  extracts key parameters, and passes them to the robot motion control component in order to appropriately adapt the motion trajectory to effectively use the tool for a task. We evaluate our approach by testing how effectively our skills can be transferred to a set of test scenarios.
\section{RELATED WORK}
\label{sec:related_work}

\begin{figure*}[t!]
	\centering
	\includegraphics[width=\textwidth]{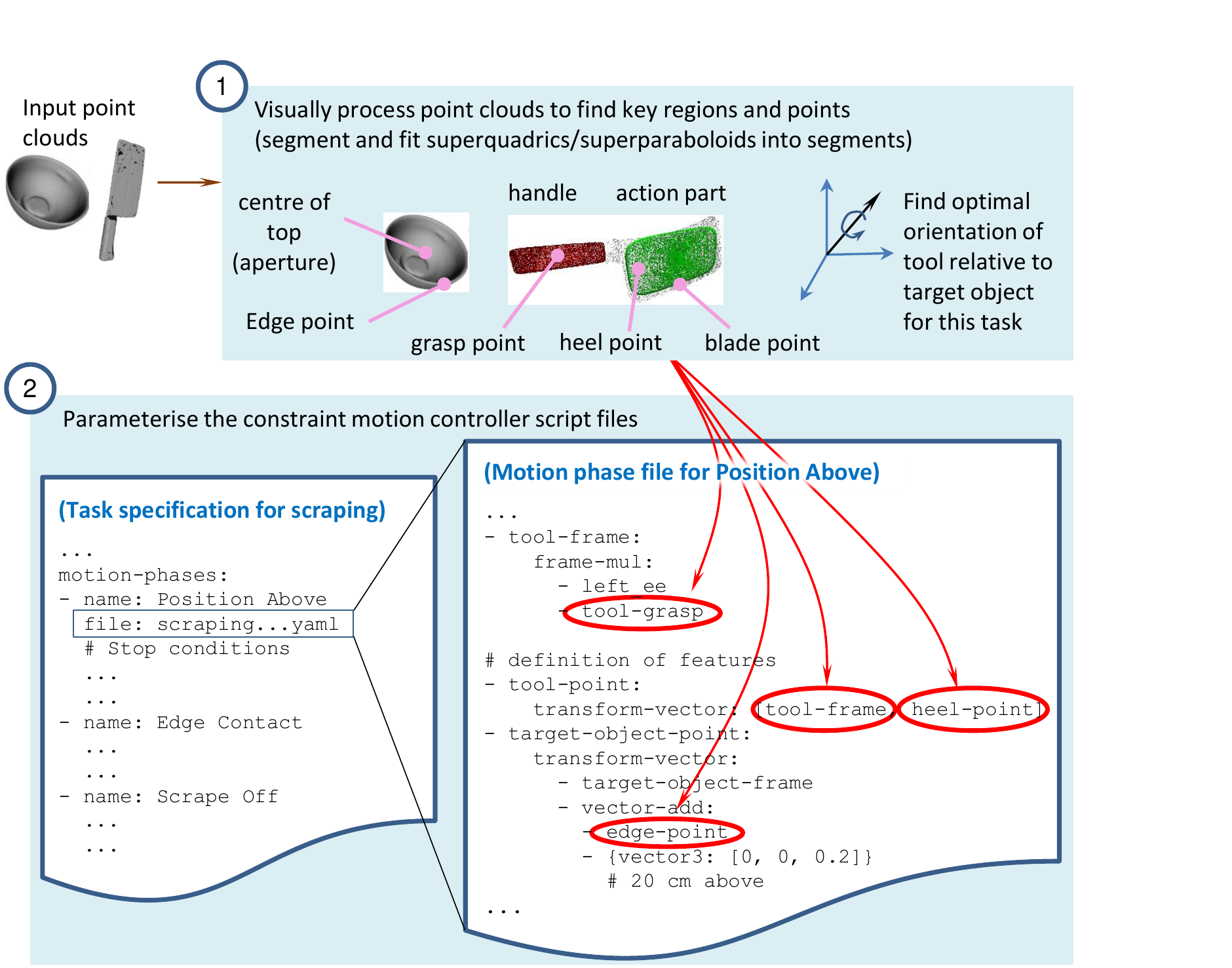}
	\caption{
	Overview of how geometric features determine motion parameters (via constraints).
}\label{overview}
\vspace{-0.5cm}
\end{figure*}

\noindent A recent  position paper on task transfer \cite{fitz2017} outlines a spectrum of increasingly difficult transfer problems depending on the level of similarity between the target scenario faced and the scenarios with which the robot is familiar. In this spectrum we tackle levels 3 and 4, i.e., replaced objects and also adjustments to the relationships between the objects during the manipulation. The next level (5) involves the introduction of new skills, such as removing a pot lid to allow scooping, which is not addressed in our work.

One branch of research related to this issue deals with tool substitution; several works
make use of some knowledge source about tools, for example a repository of objects and attributes with roles \cite{agostiniaeinszedmak2015}, or affordances modelled in description logic \cite{Awaad2013Affordance-Base}, or leveraging ConceptNet \cite{bot2015}.
We are more interested in approaches based purely on robot perception (without relying on other knowledge about the tool).
Robot vision approaches analysing affordances of tools typically learn a classifier from some visual features \cite{Myers:ICRA15,7759429}.
One of these leading approaches to affordance detection using deep learning was applied to the task of robot grasping; for this they needed to fit a minimum rectangular bounding box around the detected graspable region of an object in order to provide the required information for a grasp \cite{7759429}. Similarly we believe that fitting shapes can facilitate tool-use tasks. For example, for the container part of a spoon to be used in a manipulation, we need to identify: the orientation, the tip that should enter a liquid, the centre, etc. which is easily facilitated if we fit shapes. Our vision component \cite{AbelhaArXiv2017} does fit shapes and can identify a suitable substitute tool, and also give indications about where to grasp it and how to  orient it.
However it does not attempt to adapt a motion trajectory to accommodate a new tool.

A second branch of research deals with robot motion. In general, motion description languages are used to lift low-level task specifications to more symbolic terms.
Kresse and Beetz \cite{KresseICRA2012} can operate with high-level descriptions using terms 
like point-toward, flip-over, lift  for
describing actions.
These descriptions can be translated into low-level constraint-based movement specifications, which can be run by a motion controller with an appropriate solver.
In a similar manner, Tenorth et al. \cite{Tenorth2014}
offers a flexible way of describing
motion tasks via a library of generic motion patterns that
are composable and extensible. These patterns are then translated into specific
constraints by resolving necessary objects and relations between them, and sent to
a controller for execution. This approach allows us to assemble task descriptions
in a symbolic and reusable way, and to automatically translate them into specific constraints for a current scene.

The approach that we borrow for our motion control is described in Fang et al. \cite{DBLP:conf/iros/FangBB16}. At a high level this approach derives from Tenorth et al. \cite{Tenorth2014}, while its lower level motion control descends from  eTaSL language and eTC controller from Aertbelien and De Schutter \cite{AertbelienIROS2014}, which in turn builds on ideas such as the iTaSC framework \cite{SmitsLNEE2009}.


The solutions above provide useful tools for describing motion tasks. We want to connect this to vision, so that a robot can autonomously  parameterise its motion to adapt to a newly perceived scene.
A related approach tackles assembly tasks and uses CAD model-based vision system and an intuitive teaching interface, where a user can specify constraints between object parts \cite{7759358,7353770}. The robot then performs appropriate trajectories driven by the vision input and the constraints it needs to satisfy. This is similar to our approach except that we have an extra step in vision so that we can fit geometric shapes to new objects and do not rely on CAD models. Another related work fits geometric shapes to CAD models and recognises functional parts \cite{tenorth2013}; this was later connected with  a motion control component   \cite{Tenorth2014}.
Finally, the approach of warping point clouds from a source object to an unknown target object \cite{7041426} is a close alternative to what we propose here; the warping approach can recognise key features of the target object, such as an aperture to pour into, or a rim point where liquid exits the pouring container, and the motor program can be adapted accordingly.


\section{SYSTEM}

\noindent We start with a quick overview of the system followed by more details of its subcomponents in the subsequent subsections.
We assume the robot already has complete point cloud scans of the objects in advance, and that it has a stable grasp of both the tool and the target object. There are existing works that can solve the problem of arriving at this point \cite{KraininIJRR2011,mahler2017dex}.
For the experiments on the real robot PR2, our grasps are predetermined because there are only certain ways of grasping an object firmly by
PR2 grippers. We need a firm grasp so we can calculate object positions accurately. Also, all our tasks involve some additional force on the grasped tool, beyond its weight. Often
custom handles have to be attached to objects to achieve a reliable grasp.

The system relies on task descriptions that are hand-engineered and coded in a generic way, referring to key features of tools and target objects, such as `edge point' or `tool tip' or `centre of top'. These features can be grounded, by vision, in many different tools and target objects, thereby permitting the manipulation to be adapted to these objects.
An example of a partial task description appears at the bottom of Fig.~\ref{overview}.

In the first step (step 1 in Fig.~\ref{overview}), the computer vision module processes  the point clouds to extract key features of
objects and to determine the orientation in which the tool should be used. 
This is the creative part of the program because the system itself decides how to interpret the point cloud, in particular which part could be used for the action (scoop, cut, or scrape). The system can generate novel interpretations: e.g., for scooping, if the tool has any concave part, the system could pick this as a suitable part to use and will find the orientation which permits it to be used; similarly for scraping, any flat part could be picked as a suitable surface to scrape. The vision module is described further in Sec.~\ref{vis}.

Next the system loads the motion task descriptions, which consist of one high-level description, and one file for each phase of motion in the task (see bottom of Fig.~\ref{overview}).
These files are combined with  a template file appropriate for the robot type (e.g. PR2);
the template consists of joint and link definitions, and additional parameters. 
In step 2 in Fig.~\ref{overview}, the system
 injects the knowledge acquired about objects into the motion phase descriptions. 
Finally, it sends the prepared motion description to a controller for execution.
The robot watches the state of motion and decides that a motion phase is done when certain
specified stop conditions are met.
This process repeats until all motion phases are complete.

\begin{figure}
	\centering
	\includegraphics[width=8.5cm]{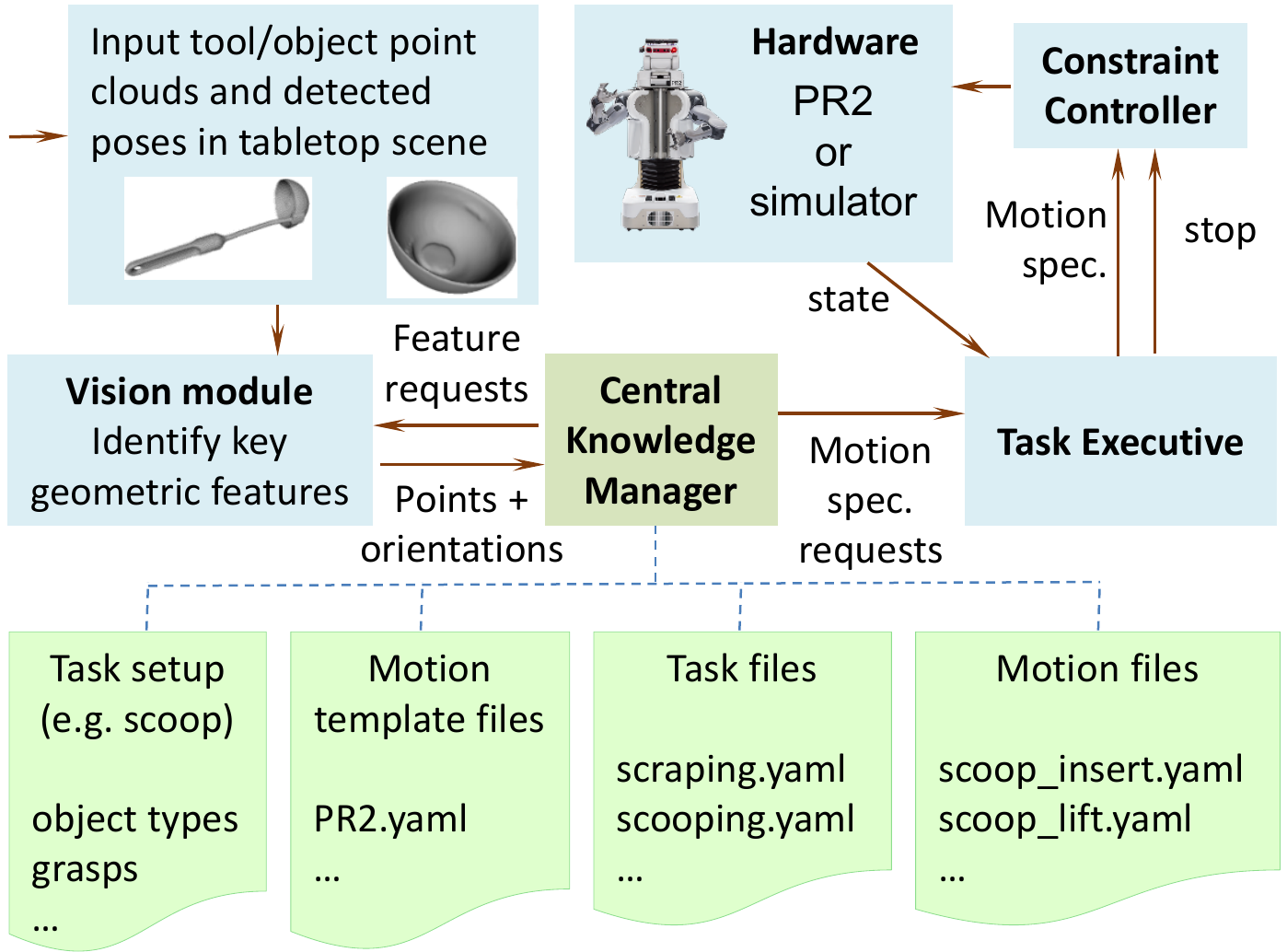}
	\caption{
	Overview of system Architecture.
}\label{architecture}
\vspace{-0.5cm}
\end{figure}

Fig.~\ref{architecture} gives an overview of the main components of the system architecture. The Vision module finds object information (edge-point, heel-point, orientation, etc.). The Central Knowledge Manager manages specifications and knowledge needed for the task. The Task Executive supervises the motion process. The
Constraint Controller uses motion control software ``Giskard''\footnote{http://giskard.de/} internally and 
translates motion description files into desired joint
velocities. 

\subsection{Vision Module}\label{vis}

\noindent This module is based on prior work \cite{AbelhaIROS2017}, which we extended for this paper. The prior work can learn the best way to use a new tool for a given task, i.e. which part of the tool should be the end effector, and what orientation it should be held in. This is learnt by simulation of different ways to use a large set of tools for five of tasks\footnote{Code is available at https://github.com/pauloabelha/enzymes/IROS2018/}. The system is model-based, using superquadrics (including superparaboloids) \cite{AbelhaSQArXiv2018} as its representation of tool parts.
The first step it performs is to segment the tools, then it fits superquadric shapes in the different segments (details in \cite{AbelhaArXiv2017}). In the current paper we used high quality full point clouds, but in our past work we also experimented with fitting in low quality and partial point clouds showing that performance was reduced but not catastrophically.
Here we extend the vision module to also identify edges of containers, and also to package it as a ROS node. The vision module can be queried in two different ways: 1) get target object information; 2) get tool information.

\subsubsection{Target object information}
\begin{figure}[t!] 
  \begin{subfigure}[b]{0.5\linewidth}
    \centering
    \includegraphics[width=0.6\linewidth]{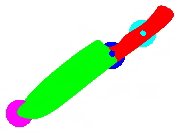} 
    \label{fig:task_sim_roll} 
    \vspace{2ex}
  \end{subfigure}
  \begin{subfigure}[b]{0.5\linewidth}
    \centering
    \includegraphics[width=0.6\linewidth]{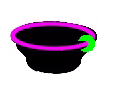} 
    \label{fig:task_sim_cut} 
    \vspace{2ex}
  \end{subfigure} 
  \vspace{-0.5cm}
  \caption{Left: Tool's outputs: heel point (blue), grasp point (cyan), tip point (magenta); Right: target object's outputs: aperture superellipse (magenta); edge point (cyan).}
  \label{fig:vision_module_outputs}
\end{figure}

The target object is assumed to be a container (modelled as superparaboloid); the information returned for it is comprised of two points: the target object's edge and top centre (which for a container is the centre of the aperture). These two points are used in all three tasks. Please see Fig.~\ref{fig:vision_module_outputs} (right) for the scraping butter task as an example

For edge detection, the module returns an edge point closest to a given point. Thus, given a point cloud of the container and an external point such as the tool's tip, the module can output the closest edge point on the container. This submodule works by first fitting a superparaboloid \cite{AbelhaArXiv2017} to the point cloud; then sampling points around its top superellipse; and choosing the one closest to a given external point. The system does not check if this point is reachable, it is assumed that it is.

\subsubsection{Tool information}
For getting tool information, the module returns grasp and end-effector points and the required final orientation of the tool relative to the target object for the task (see Fig.~\ref{fig:vision_module_outputs} right). The details returned vary depending on the task, e.g., a cut point is returned only for the cutting task.
The grasp points returned are not used in the current work because we have already fixed the robot grasp before execution starts, as explained at the start of this section. 


\subsection{Motion Control}
\noindent For each task we write configuration files for the high level task description (in terms of consisting phases of motion and stopping conditions, see  Fig.~\ref{overview} part 2, left) and for the phases of motion (in terms of constraints, see  Fig.~\ref{overview} part 2, right).
Stopping conditions take into account measured end effector velocity, desired end effector velocity or  distance from the goal. In the following the task configurations are briefly described.

 \textbf{Scraping task}: Orient the tool to point in the direction of the vertical axis through the centre of the target container; move the heel point of the tool 20cm above the edge point of the target container; move the tool down until heel contacts container edge; move the tool backwards along its major axis.

 \textbf{Scooping task}: Orient the tool to point in the direction of the container rim center; move the tool 20cm above the center; insert the tool 6.5cm below the rim center; rotate the tool horizontally while moving it 5cm towards the edge of the container; pull the tool upwards 20cm away from the rim center. We assume that the container is sufficiently deep, but ideally depth should  be automatically detected.

 \textbf{Cutting task}: Orient the tool with major axis horizontal; move the blade point of the tool 30cm above the centre point of the target; move the tool down until contact with table (distance to table goes to zero, or velocity goes to zero); move the tool backwards along its major axis.

We use Fang et al.'s \cite{DBLP:conf/iros/FangBB16} constraint based language which allows one to specify motions as a composition of various types of constraints. There are hard constraints which cannot be violated and soft constraints whose violation is minimised. 
The motion of the robot's joints is calculated by solving a minimisation problem.


\subsection{Simulation and Generating Trajectories}
\noindent 
To evaluate our system, we employ the Gazebo simulator as a core component. We simulate the PR2 robot, its environment and the actual manipulation processes. The constraint controller modules sends velocity commands to the simulated robot, and we observe the simulated action effects and contact events. The system decides when to stop one phase of motion and start the next based on two stopping conditions: Thresholds are set for both velocity and distance from goal being close to zero. These both should approach zero at the same time, triggering movement to the next motion phase; however if something unexpected happens such as collision, one will approach zero, and this will be sufficient to trigger the next motion phase. Throughout the simulation, we record the joint trajectories of the robot. On the real robot, we manually reproduce the same initial geometric setup as in the simulation and then execute the recorded trajectories from simulation. In a way, one could consider our simulation system as a physics-based planning framework.

The entire process is not real-time, but done offline and later exported to the robot. Some components are rather slow, e.g. the vision module, because it is coded in Matlab, incurs the overhead of loading Matlab Runtime. However the superquadric fitting algorithms used are fast enough to be implemented in a real-time perception system if required.







\section{EXPERIMENTAL EVALUATION}

\noindent Given a  particular tool and task, we wish to answer: Can the system adapt the motion appropriately to use this tool for the task?
We tested this for a variety of objects using a simulated environment and also a real PR2 robot. Testing in simulation used a larger set of test objects, while only a small subset was used for PR2 experiments, in order to verify that what works in simulation does indeed translate to the real robot.


\begin{figure}
	\centering
	\includegraphics[width=8cm]{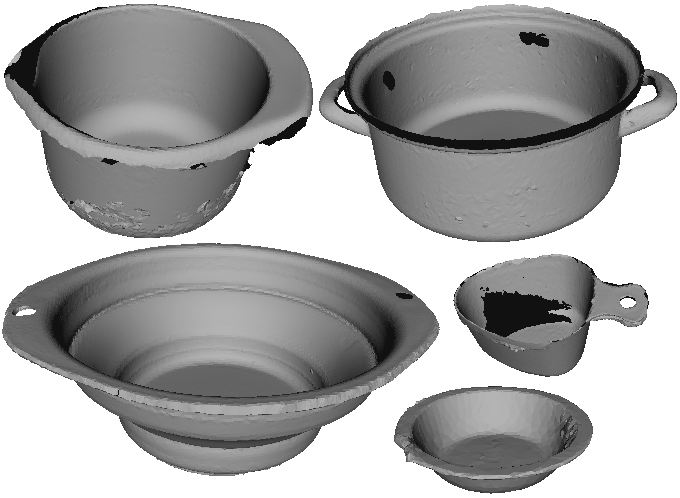}
	\caption{
	Scanned models of five containers used.
}\label{contExamples}
\end{figure}

\begin{figure}
	\centering
	\includegraphics[width=8cm]{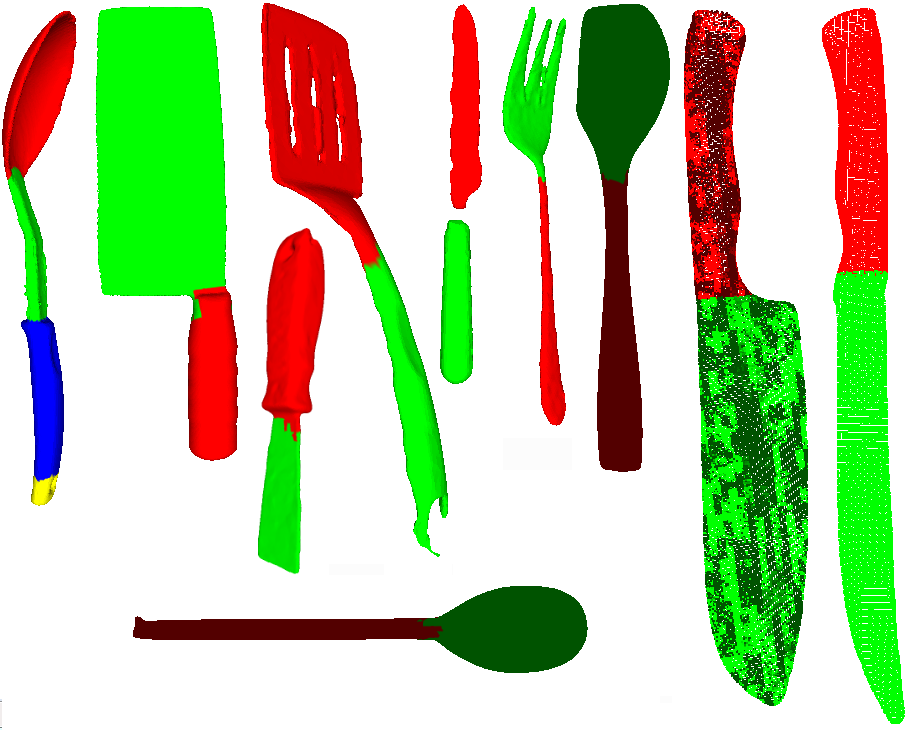}
	\caption{
	Segmented models of tools used for scraping. All are scanned models except for the two large knives on the right (constructed from CAD models).
}\label{scrapeExamples}
\end{figure}

\textbf{Scraping}: 
We tested in simulation each of the containers of Fig.~\ref{contExamples} with each of the tools of Fig.~\ref{scrapeExamples} for the scraping task, making a total of 50 tests of which 23 worked successfully (i.e. the simulated butter was scraped off and fell in the container).
The main reason for failures was the superparaboloid fitting to the containers. This is especially problematic for some of the containers which have poor point cloud scans. The rim (superellipse) of the superparaboloid is sometimes higher than the rim of the actual mesh model, meaning that the tool stops short of the container and scrapes in free space slightly above it. Furthermore the superparaboloid fitting is nondeterministic due to random planting of seeds. In the case of the 2-handled pot the superparaboloid tended to be outside the mesh and lower than the true rim. Higher quality 3D scans would fix these problems. In contrast to the containers there was no problem with the tools; all of the tools worked with at least some containers;  they were appropriately oriented  and motions adapted  to their sizes.

\begin{figure}
	\centering
	\includegraphics[width=8cm]{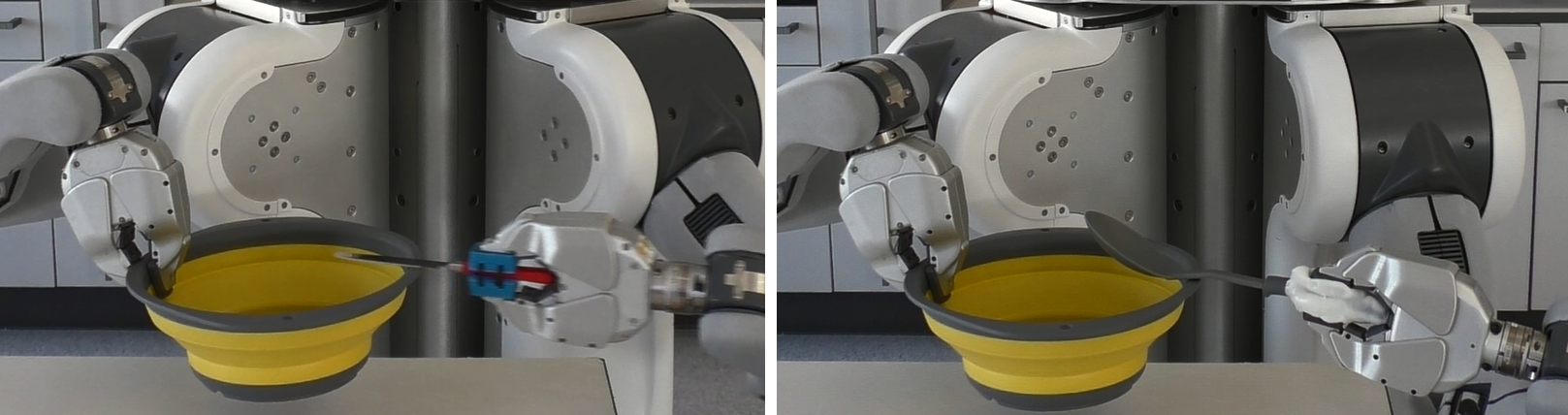}
	\caption{
	PR2 scraping across the bowl edge with a small knife (left) and large serving spoon (right). Please see the attached video for further examples on the PR2 robot.
}\label{pr2}
\end{figure}

Note that for getting the tip point of the tool the vision module does not take a point of the superquadric it has fit, but rather takes a point from the actual point cloud that is the furthest point from the grasp. A similar approach may be beneficial for the container; i.e. to find the true rim, rather than relying on the result of the superparaboloid fitting process  it could be extended upwards to the furthest extent of the point cloud.
Fig~\ref{pr2} shows the PR2 robot executing the final phase of scraping with two different tools.

\begin{figure}
	\centering
	\includegraphics[width=8cm]{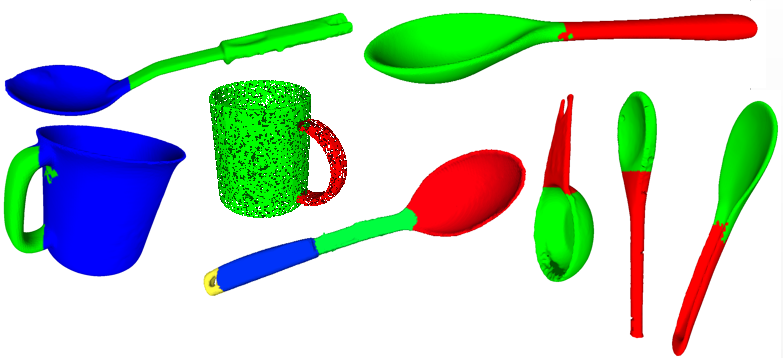}
	\caption{
	Segmented models of tools used for scooping. All are scanned models except for the central mug (constructed from CAD model).
}\label{scoop}
\end{figure}

\textbf{Scooping}: 
Scooping was tested only with tools that could be expected to be good at scooping, shown in Fig~\ref{scoop} (e.g., not knives or forks), and the containers of Fig.~\ref{contExamples}. As with the scraping experiment, containers were the limiting factor, and only the two largest containers worked for all scooping tools. Of particular note is the mug, which worked quite effectively despite being a quite different tool (see Fig.~\ref{scoop}).

\begin{figure}
	\centering
	\includegraphics[width=8cm]{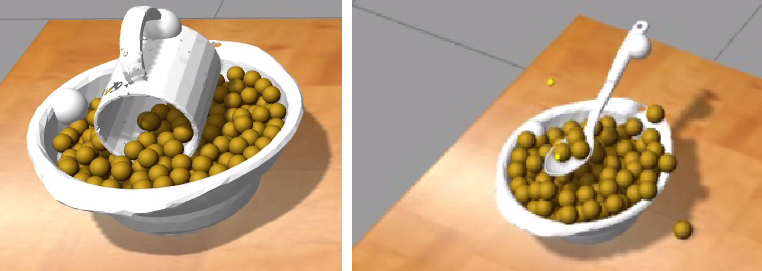}
	\caption{
	Scooping simulation with two different tools.
}\label{scooping}
\end{figure}

\textbf{Cutting}: Cutting was tested with all knives   shown in Fig~\ref{scrapeExamples} as well as the paint scraper and silicone spatula. Cutting worked correctly with all tools. There is no container object, only some simulated material to cut. The motion trajectory was adapted appropriately to the size of the tool, and the ideal orientation.

We do not have a direct comparison with a competing system using the same objects. We feel the best candidate for such a comparison would be the approach of `warped parameters' \cite{7041426} mentioned in Sec.~\ref{sec:related_work}. We select this as it was the only example we found that could deal with raw point clouds of novel objects, and adapt motions appropriately. That work was only applied to a pouring task with varied cups and bowls. If extended to our tool tasks it may be competitive with our approach.


\section{DISCUSSION AND CONCLUSIONS}

\noindent 
We have presented an integrated system combining a robot vision system with a motion control framework in order to tackle the problem of allowing tool-use manipulation skills to adapt to varied tools and target objects.

We go beyond the closest work in the literature \cite{Tenorth2014}, advancing three contributions: going from CAD models to superquadric fitting, thereby improving flexibility; considering three tasks instead of only one; and partly learning from simulation instead of relying on purely hard-coded interpretations for objects. Through these three contributions we acquire flexible  semantic grounding for our tasks.

Future work to make a more robust and complete manipulation system should perceive and monitor the states of substances (e.g. material to be scraped or scooped). Some recent works addressing this kind of perception show that tackling the perception alone is a quite challenging problem even for just one modality  \cite{7989307,7803316}. Such perception would  permit the system to react to effects during execution, which indeed is likely to be challenging, requiring time series perception data and estimation \cite{7803316}.

Currently our approach relies on the human designer identifying the key features that are needed from vision to perform tasks robustly. The human then hand codes a generic motion script for the skill, where these features are  used to parameterise the motions. In future it would  be better if the robot learns its own specific skill from demonstration in one situation, and then generalises. In fact the robot motion component that this paper is based on has already been extended in this direction \cite{DBLP:conf/iros/FangBB16}.  The next logical step would then be to tackle the full transfer problem as described by Fitzgerald et al. \cite{fitz2017}, including, e.g., when new planning steps might be introduced to deal with situations such as a pot having a lid which needs to be removed.


This system provides a high level abstraction of the key parameters in a manipulation task and therefore could be tied into cognitive robots that plan using knowledge
\cite{Tenorth2014} although we did not do that in this paper. 
An advantage that our work could bring cognitive robots is that  we can ground symbols (edge point, heel point, etc.)  in creative ways, therefore we can see the world in ways that might facilitate a plan step. We can use top down pressure from the requirements of a planning step, rather than being constrained to see the world in one way, by bottom up processing.






\section*{Acknowledgements}
This work is partially funded by:
(1) AGH University of Science and Technology, grant No 15.11.230.318.
(2) Deutsche Forschungsgemeinschaft (DFG) 
through the Collaborative Research Center 1320, \textit{EASE}.
(3) Elphinstone Scholarship from University of Aberdeen.



\addtolength{\textheight}{-9.0cm}   


\bibliographystyle{IEEEtran}
\bibliography{library,learning}




\end{document}